\documentclass[10pt,twocolumn,letterpaper]{article}

\usepackage{cvpr}              
\usepackage[accsupp]{axessibility}  
\usepackage{makecell}
\usepackage{multirow}
\usepackage{graphicx}

\usepackage[table]{xcolor}
\usepackage{xcolor}
\usepackage{appendix}

\definecolor{cvprblue}{rgb}{0.21,0.49,0.74}
\definecolor{cell}{RGB}{180, 255, 255}
\usepackage[pagebackref,breaklinks,colorlinks,allcolors=cvprblue]{hyperref}

\title{Breaking the Low-Rank Dilemma of Linear Attention}

\author{%
  Qihang Fan$^{1, 2}$, Huaibo Huang$^{1}$\thanks{Huaibo Huang is the corresponding author.}, Ran He$^{1, 2}$\\
  $^1$MAIS \& CRIPAC, Institute of Automation, Chinese Academy of Sciences, Beijing, China\\
  $^2$School of Artificial Intelligence, University of Chinese Academy of Sciences, Beijing, China\\
  \texttt{fanqihang.159@gmail.com, huaibo.huang@cripac.ia.ac.cn, }\\
  \texttt{rhe@nlpr.ia.ac.cn}
}

\begin{document}
\maketitle
\begin{abstract}
The Softmax attention mechanism in Transformer models is notoriously computationally expensive due to its quadratic complexity, posing significant challenges in vision applications. In contrast, linear attention offers a far more efficient solution by reducing the complexity to linear levels. However, linear attention often suffers significant performance degradation compared to Softmax attention. Our experiments indicate that this performance drop stems from the low-rank nature of linear attention's output feature map, which hinders its ability to adequately model complex spatial information. To address this low-rank dilemma, we conduct rank analysis from two perspectives: the KV buffer and the output features. Consequently, we introduce \textbf{Rank-Augmented Linear Attention}  (RALA), which rivals the performance of Softmax attention while maintaining linear complexity and high efficiency. Building upon RALA, we construct the \textbf{Rank-Augmented Vision Linear Transformer} (RAVLT). Extensive experiments demonstrate that RAVLT achieves excellent performance across various vision tasks. Specifically, without using any additional labels, data, or supervision during training, RAVLT achieves an \textbf{84.4\%} Top-1 accuracy on ImageNet-1k with only \textbf{26M} parameters and \textbf{4.6G} FLOPs. This result significantly surpasses previous linear attention mechanisms, fully illustrating the potential of RALA. Code will be available at \url{https://github.com/qhfan/RALA}.
\vspace{-3mm}

\end{abstract}    
\section{Introduction}
\label{sec:intro}

\begin{figure}[!ht]
    \centering
    \includegraphics[width=0.99\linewidth]{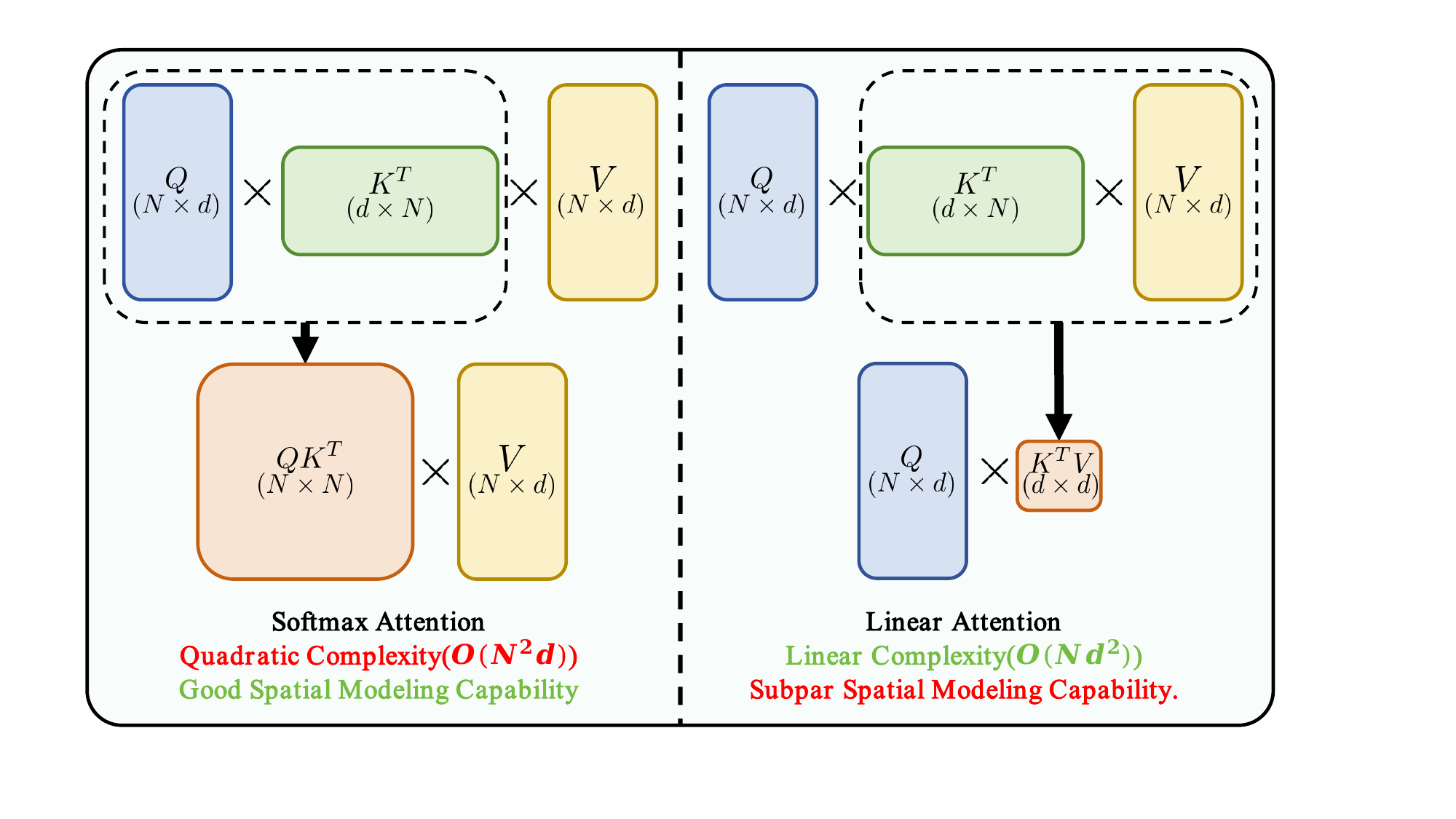}
    \vspace{-2mm}
    \caption{Comparison of Softmax attention and linear attention. Linear attention has linear complexity and high efficiency, but its spatial modeling capability is inferior to Softmax attention.}
    \vspace{-3mm}
    \label{fig:comp}
\end{figure}

In recent years, the Transformer~\cite{attention} has garnered increasing attention as a powerful foundational architecture. In the field of computer vision, Vision Transformers~\cite{vit} have made significant breakthroughs in image classification, object detection, instance segmentation, and semantic segmentation, further demonstrating the vast potential of Transformers. 

\begin{figure*}[ht]
    \centering
    \vspace{-3mm}
    \includegraphics[width=0.99\linewidth]{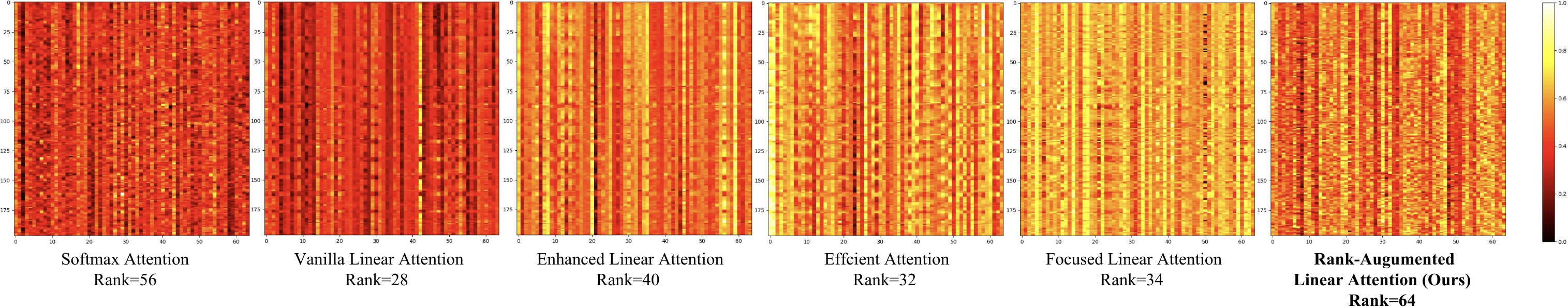}
    \vspace{-3mm}
    \caption{Comparison of feature maps output by Softmax attention and different linear attentions. All experiments are conducted based on the DeiT-T architecture, with $N=196$ and $d=64$. The full rank of matrices in the fig is 64. Compared to Softmax attention, the output features of various linear attentions exhibit significantly low-rank properties. This indicates that the diversity of features learned by linear attention is inferior to that learned by Softmax attention.}
    \vspace{-3mm}
    \label{fig:rank}
\end{figure*}

\begin{figure}[h]
    \centering
    \vspace{-3mm}
    \includegraphics[width=0.99\linewidth]{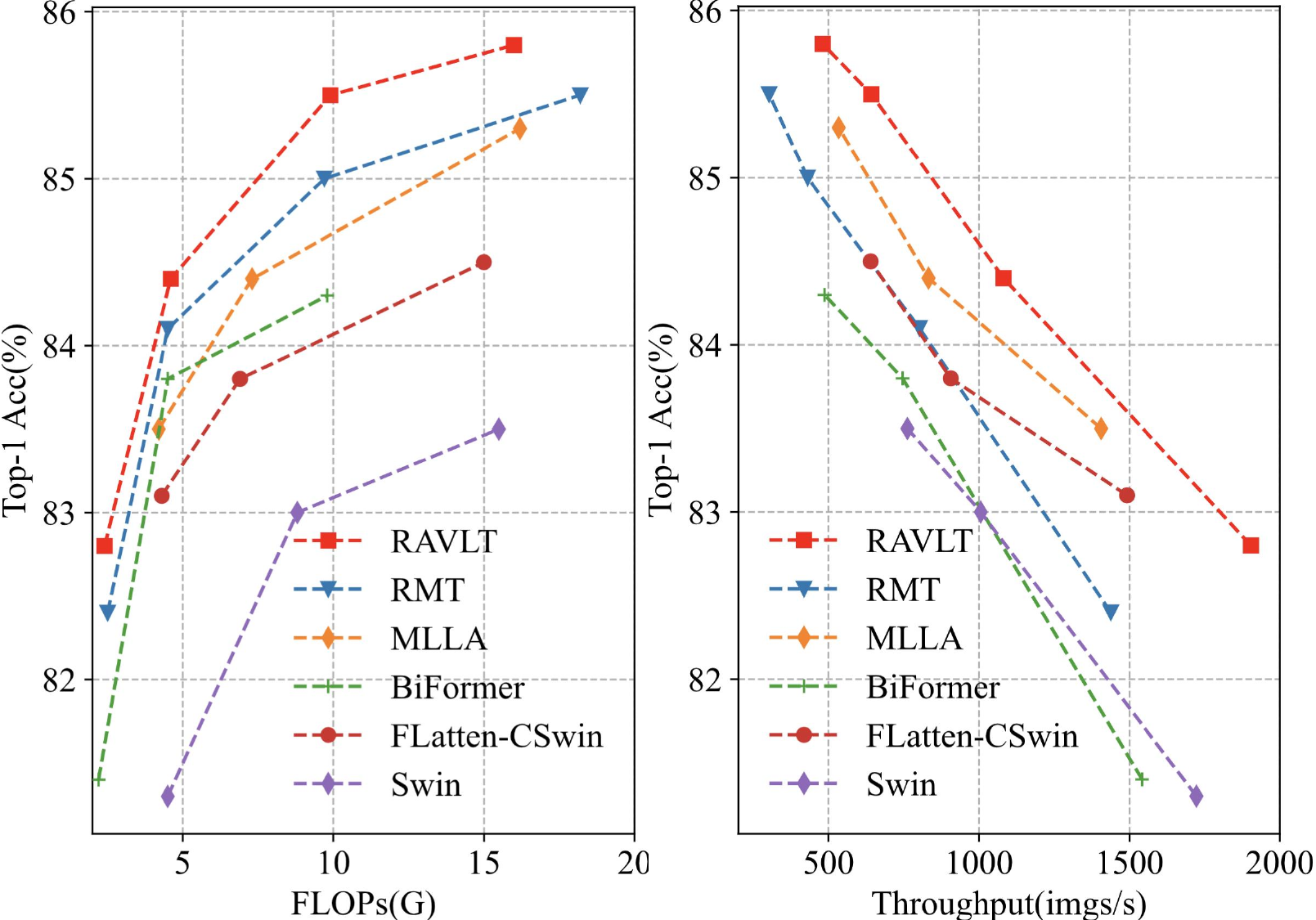}
    \vspace{-3mm}
    \caption{Comparison among models based on linear attention and Softmax attention. Our RAVLT achieves state-of-the-art results across all scales and significantly outperforms existing vision models based on linear attention.}
    \vspace{-3mm}
    \label{fig:flops-acc}
\end{figure}

However, applying Transformers to vision tasks is not straightforward. Self-attention, as the core mechanism of Transformers, has quadratic complexity. As the number of tokens increases, the computational load of self-attention grows significantly, making Transformers challenging to apply to vision tasks. Many works attempt to address this issue, with some approaches grouping vision tokens to limit the number of tokens each token can interact with~\cite{SwinTransformer, cswin, biformer, DGT, fan2024semantic, crossformer, maxvit}. While this method effectively reduces the model's complexity, it sacrifices the Transformer's crucial ability to perceive information globally. Another line of work attempts to downsample the keys and values in Softmax attention to capture global features with lower complexity~\cite{pvt,pvtv2,FAT, globalvit}. However, this approach sacrifices the model's fine-grained perception capabilities. 

Unlike the methods mentioned above, linear attention~\cite{flattentrans, MLLA, SOFT, efficientattn, linearattn, efficientvit} takes a different approach by replacing the Softmax with kernel functions and altering the computation order of $Q$, $K$, and $V$, which is shown in the Fig.~\ref{fig:comp}. This reduces the quadratic complexity of Softmax attention to linear complexity. Additionally, because its computation closely resembles Softmax attention and only uses kernel functions to approximate Softmax, linear attention possesses excellent global modeling capability. Furthermore, since linear attention does not involve downsampling, it retains the model's fine-grained perception ability.

Despite having a form and advantages very similar to Softmax attention, the actual performance of linear attention is less than satisfactory. Although it is highly efficient, its performance shows a significant gap compared to Softmax attention. To identify the cause of this performance gap, we conduct a comparison based on the DeiT-T architecture, merely replacing different attention mechanisms (vanilla linear attention, Enhanced Linear Attention~\cite{efficientvit}, Efficient Attention~\cite{efficientattn}, Focused Linear Attention~\cite{flattentrans}), with results shown in Fig.~\ref{fig:rank}. We perform rank analysis on the features of single heads from different output feature maps ($N=196$, $d=64$, with the full rank of the matrix being 64). We find that the rank of output features from existing linear attentions are significantly lower than those from Softmax attention, indicating that the diversity of features they learn is comparatively poorer.

Based on the above considerations, we aim to increase the rank of linear attention to achieve a trade-off between performance and efficiency in vision models. Specifically, we find that two computational steps in linear attention—the calculation of the KV buffer and the calculation of the output features—both affect its low-rank characteristics. Based on this, we propose \textbf{Rank-Augmented Linear Attention} (RALA). In RALA, to address the low-rank issue of the KV buffer, we use a set of context-aware rearrangement coefficients to restructure the weights of each token in the KV buffer. This approach enhances the richness and diversity of information in the KV buffer, which in turn increases the rank of the matrix. For the low-rank issue of the output features, we introduce a feature interaction strategy in the channel dimension, ensuring the output features reach a full-rank state. RALA effectively enhances the rank of the feature matrices in linear attention, enabling it to model complex spatial features with linear complexity. As shown in the Fig.~\ref{fig:rank}, the output feature matrix of RALA reaches the full rank of 64.

Utilizing RALA, we construct the \textbf{Rank-Augmented Vision Linear Transformer} (RAVLT). We conduct extensive experiments with RAVLT on image classification, object detection, instance segmentation, and semantic segmentation, and RAVLT achieves results comparable to state-of-the-art Vision Transformers. RAVLT also exhibits great trade-off between performance and efficiency. As shown in Fig.~\ref{fig:flops-acc}, without using any additional training data, labels, or supervision, our model achieves an accuracy of \textbf{84.4\%} on ImageNet with only \textbf{26M} parameters and \textbf{4.6G} FLOPs. As far as we konw, this surpasses all existing vision models. 


\section{Related works}

\paragraph{Vision Transformer.}Transformer is first employed in the field of NLP~\cite{attention} and have attracted wide interest in computer vision community~\cite{vit, ViL, dfvit, cvt}. The core operation of Transformer is the vanilla Softmax attention, which faces the challenge of quadratic complexity. Many methods have been proposed to address this challenge~\cite{NAT, cmt, dat, cloformer, DGT, shunted, fan2023rmt, davit}. Some methods group tokens to limit the receptive field of each token, thereby reducing the complexity of vanilla Softmax attention to linear~\cite{SwinTransformer, fan2024semantic, cswin, DGT, focal, quadtree, Ortho, crossformer, maxvit}. Another approach downsamples the key and value, sacrificing some fine-grained perception while preserving the Transformer's ability to perceive global information, and reduces the complexity of vanilla Softmax attention to linear~\cite{pvt, pvtv2, cmt, FAT, dat, regionvit}. Many methods also try to combine classical CNNs with Transformers, leveraging the excellent local perception capabilities of CNNs with the global perception abilities of Transformers to create a powerful vision backbone~\cite{cloformer, conv2former, hornet, cmt, VAN, focalnet, iformer, structvit, coatnet, cotnet}. All these models rely on the vanilla Softmax attention, which incurs quadratic complexity.

\paragraph{Linear Attention.}Linear attention uses kernel functions to approximate the Softmax function, allowing it to change the computation order of $Q$, $K$, and $V$ in vanilla Softmax attention, thereby reducing the computational complexity from quadratic to linear. Although the efficiency of linear attention is significantly improved compared to vanilla Softmax attention, its performance still lags considerably behind. Many methods have been proposed to improve this~\cite{SOFT, flattentrans, MLLA, xiong2021nystromformer, you2023castling, efficientattn}. Specifically, FLatten-Transformer~\cite{flattentrans} proposes focused function to improve the focus ability of linear attention. MLLA~\cite{MLLA} connects the linear attention with Mamba~\cite{mamba}. Most of other methods start by using kernel functions to approximate Softmax or introduce DWConv to increase feature diversity. However, the low-rank issue still exists, as shown in Fig.~\ref{fig:rank}.

\section{Method}
\subsection{Preliminary}
\paragraph{Softmax Attention.}Softmax attention is the classic mechanism used in Transformers~\cite{attention} and employed in ViT and its various variants~\cite{vit, SwinTransformer, cswin, stvit, structvit, fan2023rmt}. Specifically, given the input tokens $X\in \mathbb{R}^{N\times d}$, Softmax attention can generally be represented as:
\begin{equation}
    Y_i=\sum_{j=1}^N\frac{{\rm Sim}(Q(X_i), K(X_j))}{\sum_{m=1}^N {\rm Sim}(Q(X_i), K(X_m))}V(X_{j})
\end{equation}
where $i,j$ are the index of tokens, $X_i, Y_i \in \mathbb{R}^{1\times d}$. $Q(.)$, $K(.)$ and $V(.)$ are the linear transformations for $X$. For simplicity, we use $Q_i$ to refer to $Q(X_i)$, and similarly for $K(.)$ and $V(.)$. In Softmax attention, it is generally expressed as: ${\rm Sim}(Q_i, K_j)={\rm exp}(Q_iK_j^T/\sqrt{d})$. Since the computation of Softmax attention must be performed between all queries and keys, its complexity is $O(N^2d)$, exhibiting quadratic growth as the number of tokens increases. Therefore, applying it directly in vision models results in significant computational overhead, especially when the number of tokens is large.

\paragraph{Linear Attention.}Compared to Softmax attention, linear attention uses kernel functions to approximate the ${\rm Sim}(.,.)$, allowing the computation of ${\rm Sim}(Q_i, K_j)$ to be decomposed into ${\rm Sim}(Q_i, K_j)=\kappa (Q_i)\kappa(K_j)^T$. Specifically, the computation of linear attention is shown as follows:
\begin{equation}
    \begin{aligned}
    Y_i & = \sum_{j=1}^N\frac{{\rm Sim}(Q_i, K_j)}{\sum_{m=1}^N {\rm Sim}(Q_i, K_m)}V_j\\
    & = \sum_{j=1}^N\frac{\kappa(Q_i)\kappa(K_j)^T}{\sum_{m=1}^N \kappa(Q_i)\kappa(K_m)^T}V_j \\
    & = \frac{\kappa(Q_i)(\sum_{j=1}^{N}\kappa(K_j)^TV_j)}{\kappa(Q_i)(\sum_{m=1}^N \kappa(K_m)^T)}
    \end{aligned}
\end{equation}
where $\kappa(.)$ is the kernel function. Compared to the computation order $(QK^T)V$ in Softmax attention, the computation order $Q(K^TV)$ in linear attention reduces the complexity with respect to the number of tokens ($N$) from quadratic $O(N^2d)$ to linear $O(Nd^2)$. However, the reduction in complexity also results in a decline in performance. As shown in Fig.~\ref{fig:rank}, the output features of linear attention have a low rank, indicating that linear attention fails to learn diverse visual features.

\begin{figure}
    \centering
    \includegraphics[width=1.00\linewidth]{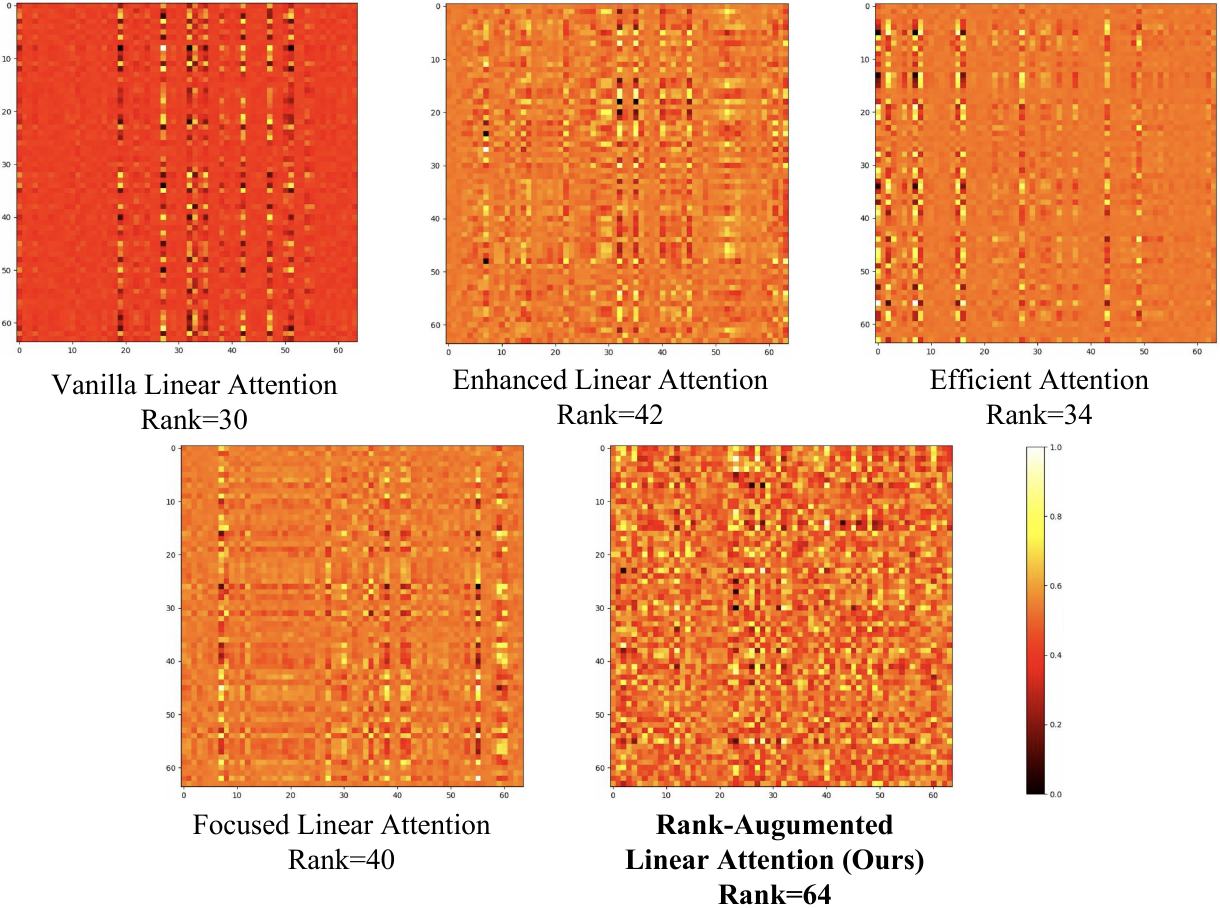}
    \vspace{-3mm}
    \caption{Visualization of the rank analysis of the KV buffer for different linear attention mechanisms. The KV buffer $\sum_{j=1}^{N}\kappa(K_j)^TV_j \in \mathbb{R}^{64\times 64}$. }
    \vspace{-3mm}
    \label{fig:rankkv}
\end{figure}

\subsection{Rank-Augmented Linear Attention}
In the general expression of linear attention, since the denominator $\kappa(Q_i)(\sum_{j=1}^N \kappa(K_j)^T) \in \mathbb{R}^{1\times 1}$, the rank of the output feature matrix $Y$ only depends on the numerator. Therefore, in the subsequent analysis, we focus solely on the numerator $\kappa(Q_i)(\sum_{j=1}^{N}\kappa(K_j)^TV_j)$.

\paragraph{Augment the rank of the KV buffer.}Each $Q_i$ needs to be multiplied by the KV buffer ($\sum_{j=1}^N\kappa (K_j)^TV_j$) to obtain the final $Y_i$. Therefore, the diversity of the features in the KV buffer, reflected by its rank, dictates the diversity of the final output features. Thus, we first analyze the rank of the KV buffer. As shown in Fig.~\ref{fig:rankkv}, the rank of the KV buffer is relatively low in almost all linear attentions. This means that the KV buffer does not store sufficient information for each $Q_i$ to query. The feature diversity in the KV buffer is poor.

To increase the rank of the KV buffer, we introduce operations into its calculation to eliminate linear dependencies between different rows and columns. Beyond the feature diversity, the rank of the KV buffer is also closely related to the amount of information it contains, increasing the information content in the KV buffer will simultaneously raise its rank. Based on this consideration, rather than assigning a weight of 1 to each $\kappa(K_j)^TV_j, j\in [1, N]$ as in other linear attentions~\cite{efficientattn, flattentrans}, we aim to highlight the significance of each token in the KV buffer by assigning higher weights to the tokens that contain more information. That is:
\begin{equation}
\begin{aligned}
    B=\sum_{j=1}^N \alpha_j \kappa(K_j)^TV_j, \quad \sum_{j=1}^N \alpha_j=N \\
\end{aligned}
\end{equation}
where $B$ denotes the KV buffer, $\alpha_j$ is the weight coefficient of the j-th token. 

To determine the value of $\alpha_j$, we draw inspiration from the design of Softmax attention. In Softmax attention, the attention scores between tokens are generally considered to represent the degree of correlation between them. Tokens more correlated with the global token are considered to contain more important information\cite{evit}. Based on this fact, by calculating the attention score of the global query to each key, we determine the proportion of each token in the KV buffer, resulting in the following:
\begin{equation}
    \begin{aligned}
        & Q_g = \frac{\sum_{i=1}^N Q_i}{N} \\
         \alpha_j & = N\times \frac{{\rm exp}(Q_g \kappa(K_j)^T)}{\sum_{m=1}^N{\rm exp}(Q_g \kappa(K_m)^T)}
    \end{aligned}
\end{equation}

As shown in Fig.~\ref{fig:rankkv}, after introducing $\alpha_j$ as the modulation coefficient for $\kappa(K_j)^TV_j$, the information content in the KV buffer increases, which significantly enhances its rank. More analysis about why the introduction of $\alpha_j$ can increase the matrix's rank can be found in \textcolor{red}{Appendix}.

\textcolor{red}{It should be emphasized that the summation of matrices can indeed increase their rank.} This is fundamentally different from a simple linear combination, which cannot increase the rank. For example:
\[
1 \times 
\begin{bmatrix} 
1 & 1 & 1 \\ 
2 & 2 & 2 \\ 
3 & 3 & 3 
\end{bmatrix} 
+ 
1 \times 
\begin{bmatrix} 
1 & 2 & 3 \\ 
1 & 2 & 3 \\ 
1 & 2 & 3 
\end{bmatrix} 
= 
\begin{bmatrix} 
2 & 3 & 4 \\ 
3 & 4 & 5 \\ 
4 & 5 & 6 
\end{bmatrix}
\]
A linear combination of two rank-1 matrices can result in a matrix with rank 3.

\vspace{-5mm}

\begin{figure}
    \centering
    \includegraphics[width=0.99\linewidth]{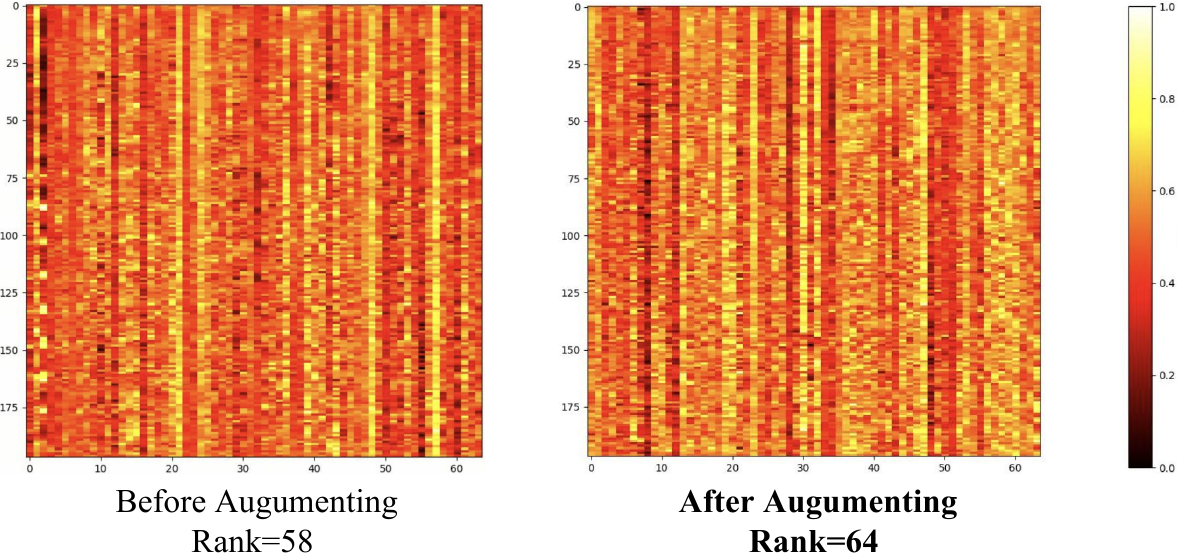}
    \vspace{-2mm}
    \caption{Visualization of the output features' ($Y\in \mathbb{R}^{N\times d}$, $N=196$, $d=64$) rank analysis.}
    \vspace{-3mm}
    \label{fig:outrank}
\end{figure}

\paragraph{Augment the rank of the output features.} In the fundamentals of matrix theory, for matrices $C$ and $D$, we have: ${\rm Rank}(CD) \leq {\rm min}({\rm Rank}(C),\:{\rm Rank}(D))$. Therefore, when calculating the output features $Y=\kappa(Q)B$, its rank:
\begin{equation}
\begin{aligned}
    & {\rm Rank}(Y) \leq {\rm min}({\rm Rank}(\kappa(Q)), \:{\rm Rank}(B))\leq d\\
    Y\in &\mathbb{R}^{N\times d},\: \kappa(Q)\in \mathbb{R}^{N\times d},\:B=\sum_{j=1}^N \alpha_j \kappa(K_j)^TV_j\in \mathbb{R}^{d\times d}
\end{aligned}
\end{equation}
This indicates that even though the KV buffer $B$ reaches a full-rank state, the output feature matrix is still highly likely to be in a low-rank state (${\rm Rank}(Y)<d$), meaning information loss occurs when computing each $Y_i$. As shown in Fig.~\ref{fig:outrank}, we confirm this point when performing rank analysis on the output features. The output features exhibit a certain degree of rank reduction.
To restore the output feature matrix $Y$ to a full-rank state, we aim to complete the information loss for each $Y_i$.

\begin{figure}[t]
    \centering
    \includegraphics[width=0.95\linewidth]{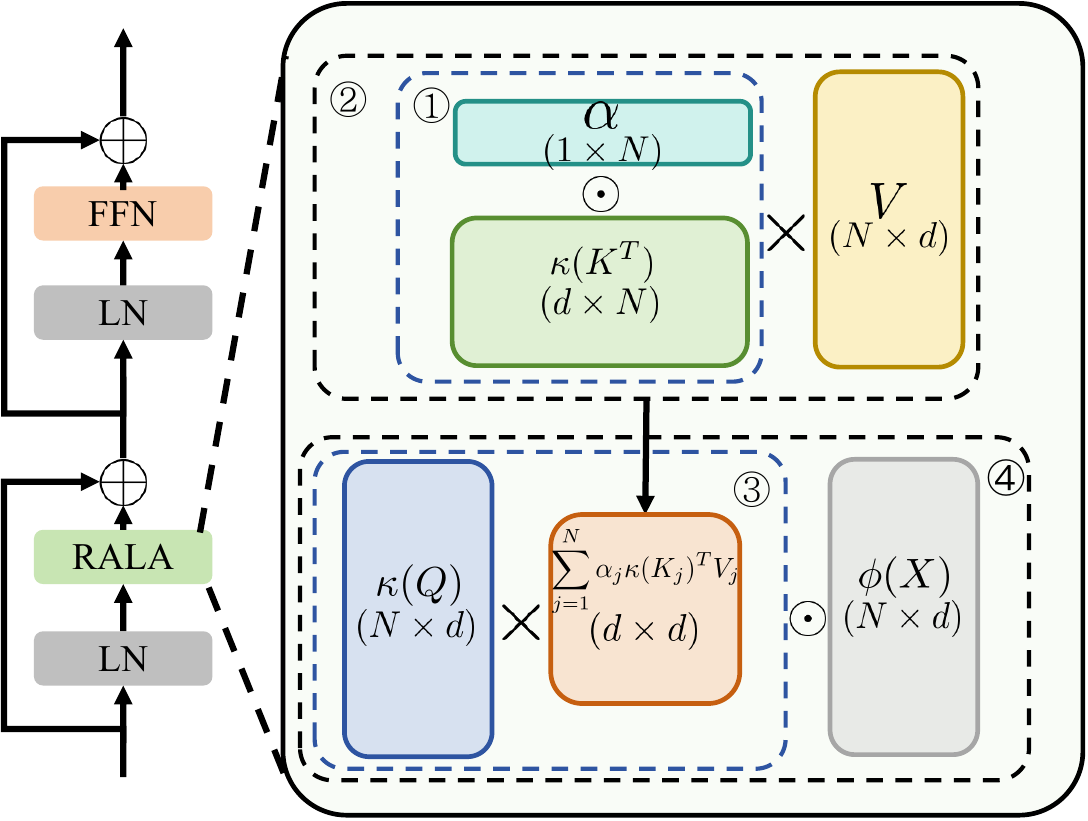}
    \vspace{-2mm}
    \caption{RAVLT's block (left) and RALA (right). We employ serial numbers to represent the sequence of calculations. $\odot$ represents Hadamard product, and $\times$ represents matrix multiplication.}
    \vspace{-3mm}
    \label{fig:RALA}
\end{figure}

Specifically, we post-process each output feature $Y_i$ by modulating $Y_i$ with a transformed form that contains the information of the original token $X_i$. At this stage, $Y_i$ has already completed token mixing, so this modulation is applied only along the channel dimension. Consequently, each output feature becomes:
\begin{equation}
    Y_i=\phi (X_i) \odot(\kappa(Q_i)\sum_{j=1}^N \alpha_j\kappa(K_j)^TV_j)
\end{equation}
where $\odot$ denotes the Hadamard product. As shown in the Fig.~\ref{fig:outrank}, After introducing the modulation coefficients, the feature diversity of the output feature matrix is significantly enhanced, transforming it from a low-rank state to a full-rank state. This indicates that token-specific modulation adds a degree of feature diversity to the output features. More analysis can be found in the \textcolor{red}{Appendix}.

\paragraph{Design of the block.}Fig.~\ref{fig:RALA} illustrates the design of the entire block. Our block design follows the classic Transformer block design~\cite{attention}, consisting of linear attention and FFN. All the networks in this paper are constructed by stacking this basic block. We use a simple linear projection to serve as the transformation function $\phi(.)$. As for $\kappa(.)$, we have $\kappa(.)={\rm Elu}(.)+1$. For each block, we use CPE~\cite{CPVT} to serve as the positional encoding. The CPE is a simple $3\times 3$ depth-wise convolution with residual connection.

\subsection{Implementation Details}
\begin{table}[t]
    \centering
    \setlength{\tabcolsep}{1.6mm}
    \scalebox{0.78}{
    \begin{tabular}{c|c c c|c c}
    \toprule[1pt]
    Model & Blocks & Channels & Heads & \makecell{Params\\(M)} & \makecell{FLOPs\\(G)} \\
    \midrule[0.5pt]
    RAVLT-T & [2,2,6,2] & [64,128,256,512] & [1,2,4,8] & 15 & 2.4 \\
    RAVLT-S & [3,5,9,3] & [64,128,320,512] & [1,2,5,8] & 26 & 4.6 \\
    RAVLT-B & [4,6,12,6] & [96,192,384,512] & [1,2,6,8] & 48 & 9.9 \\
    RAVLT-L & [4,7,19,8] & [96,192,448,640] & [1,2,7,10] & 95 & 16.0 \\
    \bottomrule[1pt]
    \end{tabular}}
    \vspace{-3mm}
    \caption{Architecture variants of RAVLT. The FLOPs are measured at resolution $224\times 224$.}
    \vspace{-3mm}
    \label{tab:av}
\end{table}
Based on the block in the Fig.~\ref{fig:RALA}, we construct the \textbf{Rank-Augmented Vision Linear Transformer} (RAVLT), a hierarchical general vision backbone. As shown in the Tab.~\ref{tab:av}, following the previous works~\cite{fan2023rmt, biformer, stvit}, we build three RAVLT backbones with different settings of block number and channel number in each stage. In each stage, the image downsampling rates are $\frac{1}{4}$, $\frac{1}{8}$, $\frac{1}{16}$, and $\frac{1}{32}$, respectively. At the beginning of each stage, we use a $3\times3$ convolution with the stride of $2$ to downsample the image.

\section{Experiments}

We conduct extensive experiments on multiple vision tasks, such as image classification, object detection, instance segmentation, and semantic segmentation. We also make ablation studies to validate the importance of each components in RALA. \textcolor{red}{More details can be found in Appendix}. 

\subsection{Image Classification}
\paragraph{Settings.}We train our models on ImageNet-1K~\cite{imagenet} from scratch and follow the same training strategy in DeiT~\cite{deit} with the only supervision being classification loss.  The maximum rates of increasing stochastic depth~\cite{droppath} are set to 0.1/0.15/0.4/0.55 for RAVLT-T/S/B/L, respectively. We evaluate the models on ImageNet~\cite{imagenet} and ImageNet-V2~\cite{imagenetv2}. 

\begin{figure}[ht]
    \centering
    \includegraphics[width=0.95\linewidth]{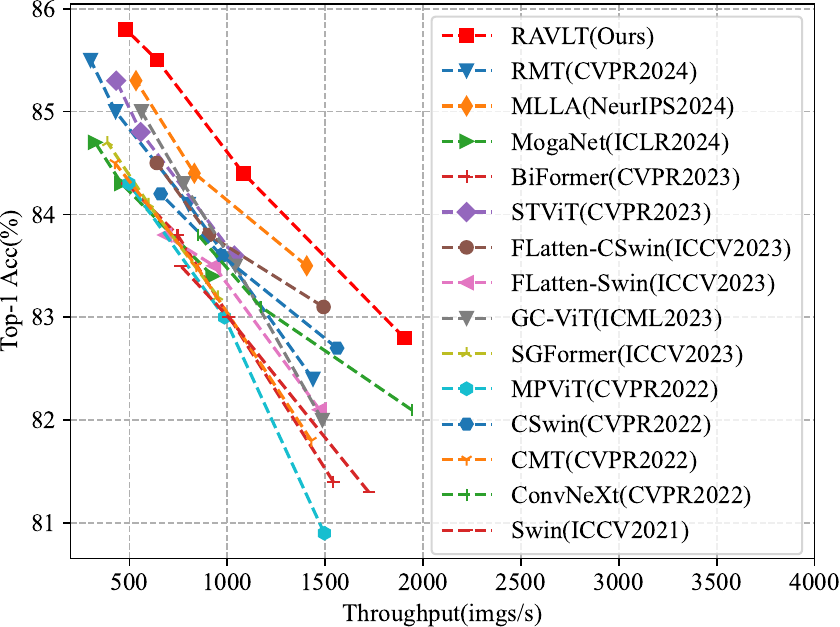}
    \vspace{-3mm}
    \caption{Comparison of multiple general backbones' inference speed. The speed are measured on A100. RAVLT exhibits the best trade-off between efficiency and performance.}
    \vspace{-3mm}
    \label{fig:effcomp}
\end{figure}

\begin{table}[ht]
    \setlength{\tabcolsep}{1.1mm}
    \centering
    \scalebox{0.78}{
    \begin{tabular}{c|c|c c|c}
    \toprule[1pt]
         Cost & Model & \makecell{Params(M)} & \makecell{FLOPs(G)} &\makecell{Top1-acc(\%)}\\
         \midrule[0.5pt]
         \multirow{5}{*}{\rotatebox{90}{\makecell{Tiny model\\$\sim 2.5$G}}}
         & tiny-MOAT-2~\cite{MOAT} & 10 & 2.3  & 70.1 \\
         & BiFormer-T~\cite{biformer} & 13 & 2.2 & 70.7 \\
         & SMT-T~\cite{SMT} & 12 & 2.4 & 71.0 \\
         & RMT-T~\cite{fan2023rmt} & 14 & 2.5 & 72.1 \\
         & \cellcolor{cell}RAVLT-T & \cellcolor{cell}15 & \cellcolor{cell}2.4 & \cellcolor{cell}\textbf{72.7} \\
         \midrule[0.5pt]
         \multirow{7}{*}{\rotatebox{90}{\makecell{Small model\\$\sim 4.5$G}}}
         & DeiT-S & 22 & 4.6 & 68.5 \\
         & MOAT-0~\cite{MOAT} & 28 & 5.7 & 72.8 \\
         & SMT-S~\cite{SMT} & 21 & 4.7 & 73.3 \\
         & MLLA-T~\cite{MLLA} & 25 & 4.2 & 73.3 \\
         & BiFormer-S~\cite{biformer} & 26 & 4.5 & 73.6 \\
         & RMT-S~\cite{fan2023rmt} & 27 & 4.5 & 74.1 \\
         & \cellcolor{cell}RAVLT-S & \cellcolor{cell}26 & \cellcolor{cell}4.6 & \cellcolor{cell}\textbf{74.9} \\
         \midrule[0.5pt]
         \multirow{6}{*}{\rotatebox{90}{\makecell{Base model\\$\sim 10.0$G}}}
         & XCiT-S24~\cite{xcit} & 48 & 9.1 & 73.3 \\
         & BiFormer-B~\cite{biformer} & 57 & 9.8 & 74.0 \\
         & MOAT-1~\cite{MOAT} & 42 & 9.1 & 74.2 \\
         & MLLA-S~\cite{MLLA} & 43 & 7.3 & 74.9 \\
         & RMT-B~\cite{fan2023rmt} & 54 & 9.7 & 75.6 \\
         & \cellcolor{cell}RAVLT-B & \cellcolor{cell}48 & \cellcolor{cell}9.9 & \cellcolor{cell}\textbf{76.3} \\
         \midrule[0.5pt]
         \multirow{5}{*}{\rotatebox{90}{\makecell{Large model\\$\sim 15.0$G}}}
         & DeiT-B~\cite{deit} & 86 & 17.5 & 71.5 \\
         & MOAT-2~\cite{MOAT} & 73 & 17.2 & 74.3 \\
         & RMT-L~\cite{fan2023rmt} & 95 & 18.2 & 76.3 \\
         & MLLA-B~\cite{MLLA} & 96 & 16.2 & 76.7 \\
         & \cellcolor{cell}RAVLT-L & \cellcolor{cell}95 & \cellcolor{cell}16.0 & \cellcolor{cell}\textbf{76.8} \\
    \bottomrule[1pt]
    \end{tabular}}
    \vspace{-3mm}
    \caption{Classification results of different general backbones on ImageNet-V2}
    \vspace{-5mm}
    \label{tab:INV2}
\end{table}

\begin{table*}[ht]
    \centering
    \setlength{\tabcolsep}{1.8mm}
    \subfloat{
    \scalebox{0.78}{
    \begin{tabular}{c|c|c|c c|c}
    \toprule[1pt]
         Cost & Model & Type & \makecell{Parmas\\(M)} & \makecell{FLOPs\\(G)} & \makecell{Top1-acc\\(\%)}\\
         \midrule[0.5pt]
         \multirow{11}{*}{\rotatebox{90}{\makecell{Tiny model\\$\sim 2.5$G}}} 
         & VAN-B1~\cite{VAN} & CNN & 14 & 2.5 & 81.1 \\
         & Conv2Former-N~\cite{conv2former} & CNN & 15 & 2.2 & 81.5 \\
         \cline{2-6}
         & MPViT-XS~\cite{mpvit} & Trans & 11 & 2.9 & 80.9 \\
         & NAT-M~\cite{NAT} & Trans & 20 & 2.7 & 81.8 \\ 
         & GC-ViT-XT~\cite{globalvit} & Trans & 20 & 2.6 & 82.0 \\
         & SMT-T~\cite{SMT} & Trans &12 & 2.4 & 82.2 \\
         & RMT-T~\cite{fan2023rmt} & Trans & 14 & 2.5 & 82.4 \\
         \cline{2-6}
         & FL-PVT-T~\cite{flattentrans} & Linear & 12 & 2.0 & 77.8 \\
         & SOFT-Tiny~\cite{SOFT} & Linear & 13 & 1.9 & 79.3 \\
         & FL-PVTv2-B1~\cite{flattentrans} & Linear & 13 & 2.2 & 79.5 \\
         & \cellcolor{cell}RAVLT-T & \cellcolor{cell}Linear & \cellcolor{cell}15 & \cellcolor{cell}2.4 & \cellcolor{cell}\textbf{82.8} \\
         \midrule[1pt]
         \multirow{15}{*}{\rotatebox{90}{\makecell{Small model\\$\sim 4.5$G}}}
         & ConvNeXT-T~\cite{convnext} & CNN & 29 & 4.5 & 82.1 \\
         & MogaNet-S~\cite{iclr2024MogaNet} & CNN & 25 & 5.0 & 83.4 \\
         & InternImage-T~\cite{internimage} & CNN & 30 & 5.0 & 83.5 \\
         \cline{2-6}
         & Swin-T~\cite{SwinTransformer} & Trans & 29 & 4.5 & 81.3 \\
         & StructViT-S-8-1~\cite{structvit} & Trans & 24 & 5.4 & 83.3 \\
         & MOAT-0~\cite{MOAT} & Trans & 28 & 5.7 & 83.3 \\
         & FAT-B3~\cite{FAT} & Trans & 29 & 4.4 & 83.6 \\
         & SMT-S~\cite{SMT} & Trans & 20 & 4.8 & 83.7 \\
         & BiFormer-S~\cite{biformer} & Trans & 26 & 4.5 & 83.8 \\
         & RMT-S~\cite{fan2023rmt} & Trans & 27 & 4.5 & 84.1 \\
         \cline{2-6}
         & FL-Swin-T~\cite{flattentrans} & Linear & 29 & 4.5 & 82.1 \\
         & SOFT-Small~\cite{SOFT} & Linear & 24 & 3.3 & 82.2 \\
         & FL-CSwin-T~\cite{flattentrans} & Linear & 21 & 4.3 & 83.1 \\
         & MLLA-T~\cite{MLLA} & Linear & 25 & 4.2 & 83.5 \\
         & \cellcolor{cell}RAVLT-S & \cellcolor{cell}Linear & \cellcolor{cell}26 & \cellcolor{cell}4.6 & \cellcolor{cell}\textbf{84.4} \\
         \bottomrule[1pt]
    \end{tabular}}
    }
    \subfloat{
    \scalebox{0.78}{
    \begin{tabular}{c|c|c|c c|c}
    \toprule[1pt]
         Cost & Model & Type & \makecell{Parmas\\(M)} & \makecell{FLOPs\\(G)} & \makecell{Top1-acc\\(\%)}\\
         \midrule[0.5pt]
         \multirow{14}{*}{\rotatebox{90}{\makecell{Base model\\$\sim 10.0$G}}}
         & ConvNeXT-S~\cite{convnext} & CNN & 50 & 8.7 & 83.1 \\
         & InternImage-S~\cite{internimage} & CNN & 50 & 8.0 & 84.2 \\
         \cline{2-6}
         & DaViT-S~\cite{davit} & Trans & 50 & 8.8 & 84.2 \\
         & StructViT-B-8-1~\cite{structvit} & Trans & 52 & 12.0 & 84.3 \\
         & BiFormer-B~\cite{biformer} & Trans & 57 & 9.8 & 84.3 \\
         & GC-ViT-S~\cite{globalvit} & Trans & 51 & 8.5 & 84.3 \\
         & STViT-B~\cite{stvit} & Trans & 52 & 9.9 & 84.8 \\
         & RMT-B~\cite{fan2023rmt} & Trans & 54 & 9.7 & 85.0 \\
         \cline{2-6}
         & SOFT-Large~\cite{SOFT} & Linear & 64 & 11.0 & 83.1 \\
         & FL-Swin-S~\cite{flattentrans} & Linear & 51 & 8.7 & 83.5 \\
         & FL-CSwin-S~\cite{flattentrans} & Linear & 35 & 6.9 & 83.8 \\
         & MLLA-S~\cite{MLLA} & Linear & 43 & 7.3 & 84.4 \\
         & \cellcolor{cell}RAVLT-B & \cellcolor{cell}Linear & \cellcolor{cell}48 & \cellcolor{cell}9.9 & \cellcolor{cell}\textbf{85.5} \\
         \midrule[1pt]
         \multirow{14}{*}{\rotatebox{90}{\makecell{Large model\\$\sim 15.0$G}}}
         & HorNet-B~\cite{hornet} & CNN & 88 & 15.5 & 84.3 \\
         & InterImage-B~\cite{internimage} & CNN & 97 & 16.0 & 84.9 \\
         \cline{2-6}
         & SMT-L~\cite{SMT} & Trans & 81 & 17.7 & 84.6 \\
         & SG-Former-B~\cite{sgformer} & Trans & 78 & 15.6 & 84.7 \\
         & GC-ViT-B~\cite{globalvit} & Trans & 90 & 14.8 & 85.0 \\
         & FAT-B5~\cite{FAT} & Trans & 88 & 15.1 & 85.2 \\
         & STViT-L~\cite{stvit} & Trans & 95 & 15.6 & 85.3 \\
         & RMT-L~\cite{fan2023rmt} & Trans & 95 & 18.2 & 85.5 \\
         \cline{2-6}
         & SOFT-Huge~\cite{SOFT} & Linear & 87 & 16.3 & 83.3 \\
         & FLatten-Swin-B~\cite{flattentrans} & Linear & 89 & 15.4 & 83.8 \\
         & FLatten-CSwin-B~\cite{flattentrans} & Linear & 75 & 15.0 & 84.5 \\
         & MLLA-B~\cite{MLLA} & Linear & 96 & 16.2 & 85.3 \\
         & \cellcolor{cell}RAVLT-L & \cellcolor{cell}Linear & \cellcolor{cell}95 & \cellcolor{cell}16.0 & \cellcolor{cell}\textbf{85.8} \\
         \bottomrule[1pt]
    \end{tabular}
    }}
    \vspace{-3mm}
    \caption{Comparison with the state-of-the-art on ImageNet-1K classification. We use "CNN" to refer to convolutional neural networks, "Trans" to refer to Vision Transformers, and "Linear" to refer to models based on linear operators. }
    \vspace{-5mm}
    \label{tab:IN1K}
\end{table*}

\vspace{-3mm}

\paragraph{Results.}In Tab.~\ref{tab:IN1K} and Tab.~\ref{tab:INV2}, we compare the performance of various models on ImageNet and ImageNet-v2, respectively. Under models of comparable size, RAVLT achieves the best results. Specifically, RAVLT-B surpasses RMT-B by \textbf{0.5\%} with the same FLOPs. This result demonstrates the superior of RAVLT.

\begin{table}[t]
    \centering
    \setlength{\tabcolsep}{0.22mm}
    \scalebox{0.75}{
    \begin{tabular}{c|c|c c|c c c c c c}
        \toprule[1pt]
         \multirow{2}{*}{Backbone} & Type & \multirow{2}{*}{\makecell{Params\\(M)}} & \multirow{2}{*}{\makecell{FLOPs\\(G)}} & \multicolumn{6}{c}{Mask R-CNN $3\times$+MS}\\
          & & & & $AP^b$ & $AP^b_{50}$ & $AP^b_{75}$ & $AP^m$ & $AP^m_{50}$ & $AP^m_{75}$\\
          \midrule[0.5pt]
          NAT-T~\cite{NAT} & Trans & 48 & 258 & 47.8 & 69.0 & 52.6 & 42.6 & 66.0 & 45.9 \\
          SMT-S~\cite{SMT} & Trans & 40 & 265 & 49.0 & 70.1 & 53.4 & 43.4 & 67.3 & 46.7\\
          RMT-S~\cite{fan2023rmt} & Trans & 46 & 262 & 50.7 & 71.9 & 55.6 & 44.9 & 69.1 & 48.4\\
          FL-Swin-T~\cite{flattentrans} & Linear & 49 & 268 & 46.5 & 68.5 & 50.8 & 42.1 & 65.4 & 45.1 \\
          MLLA-T~\cite{MLLA} & Linear & 44 & 255 & 48.8 & 71.0 & 53.6 & 43.8 & 68.0 & 46.8 \\
          \rowcolor{cell}RAVLT-S & Linear & 44 & 262 & \textbf{51.4} & \textbf{72.3} & \textbf{56.5} & \textbf{45.5} & \textbf{69.7} & \textbf{48.8} \\
          \midrule[1pt]
          InternImage-S~\cite{internimage} & CNN & 69 & 340 & 49.7 & 71.1 & 54.5 & 44.5 & 68.5 & 47.8 \\
          NAT-S~\cite{NAT} & Trans & 70 & 330 & 48.4 & 69.8 & 53.2 & 43.2 & 66.9 & 46.4 \\
          SMT-B~\cite{SMT} & Trans & 52 & 328 & 49.8 & 71.0 & 54.4 & 44.0 & 68.0 & 47.3\\
          RMT-B~\cite{fan2023rmt} & Trans & 73 & 373 & 52.2 & 72.9 & 57.0 & 46.1 & 70.4 & 49.9  \\
          MLLA-S~\cite{MLLA} & Linear & 63 & 319 & 50.5 & 71.8 & 55.2 & 44.9 & 69.1 & 48.2 \\
          \rowcolor{cell}RAVLT-B & Linear & 67 & 372 & \textbf{52.7} & \textbf{73.5} & \textbf{57.7} & \textbf{46.4} & \textbf{70.6} & \textbf{50.2} \\
          \midrule[1pt]
          InternImage-B~\cite{internimage} & CNN & 115 & 501 & 50.3 & 71.4 & 55.3 & 44.8 & 68.7 & 48.0 \\
          Swin-B~\cite{SwinTransformer} & Trans & 107 & 496 & 48.6 & 70.0 & 53.4 & 43.3 & 67.1 & 46.7 \\
          CSwin-B~\cite{cswin} & Trans & 97 & 526 & 50.8 & 72.1 & 55.8 & 44.9 & 69.1 & 48.3 \\
          \rowcolor{cell}RAVLT-L & Linear & 114 & 501 & \textbf{53.6} & \textbf{74.4} & \textbf{58.9} & \textbf{47.3} & \textbf{71.6} & \textbf{51.2} \\
          \bottomrule[1pt]
    \end{tabular}}
    \vspace{-3mm}
    \caption{Comparison to other backbones using Mask R-CNN with "$3\times+\mathrm{MS}$" schedule.}
    \vspace{-4mm}
    \label{tab:COCO3x}
\end{table}

\subsection{Inference Speed}
We evaluate the RAVLT's inference efficiency on A100, the results are shown in the Fig.~\ref{fig:effcomp}. It can be seen that our models exhibit better trade-off between performance and efficiency than other models.

\subsection{Object Detection and Instance Segmentation}
\paragraph{Settings.}We adopt MMDetection~\cite{mmdetection} to implement RetinaNet~\cite{retinanet}, Mask-RCNN~\cite{maskrcnn} and Cascade Mask R-CNN~\cite{cai18cascadercnn}. Following previous works~\cite{fan2023rmt, SwinTransformer, pvt}, we use the commonly used ``$1\times$" (12 training epochs) setting for the RetinaNet and Mask R-CNN. Besides, we use ``$3\times+\mathrm{MS}$" for Mask R-CNN and Cascade Mask R-CNN.

\begin{table}[t]
    \centering
    \setlength{\tabcolsep}{0.22mm}
    \scalebox{0.75}{
    \begin{tabular}{c|c|c c|c c c c c c}
        \toprule[1pt]
         \multirow{2}{*}{Backbone} & Type & \multirow{2}{*}{\makecell{Params\\(M)}} & \multirow{2}{*}{\makecell{FLOPs\\(G)}} & \multicolumn{6}{c}{Cascade Mask R-CNN $3\times$+MS}\\
          & & & & $AP^b$ & $AP^b_{50}$ & $AP^b_{75}$ & $AP^m$ & $AP^m_{50}$ & $AP^m_{75}$\\
          \midrule[0.5pt]
          HorNet-T~\cite{hornet} & CNN & 80 & 728 & 52.4 & 71.6 & 56.8 & 45.6 & 69.1 & 49.6 \\
          SMT-S~\cite{SMT} & Trans & 78 & 744 & 51.9 & 70.5 & 56.3 & 44.7 & 67.8 & 48.6 \\
          UniFormer-S~\cite{uniformer} & Trans & 79 & 747 & 52.1 & 71.1 & 56.6 & 45.2 & 68.3 & 48.9 \\
          RMT-S~\cite{fan2023rmt} & Trans & 83 & 741 & 53.2 & 72.0 & 57.8 & 46.1 & 69.8 & 49.8\\
          FL-Swin-T~\cite{flattentrans} & Linear & 87 & 747 & 50.8 & 69.6 & 55.1 & 44.1 & 67.0 & 48.1 \\
          \rowcolor{cell}RAVLT-S & Linear & 82 & 741 & \textbf{54.2} & \textbf{72.9} & \textbf{58.7} & \textbf{46.8} & \textbf{70.5} & \textbf{50.9} \\
          \midrule[1pt]
          HorNet-S~\cite{hornet} & CNN & 108 & 827 & 53.3 & 72.3 & 57.8 & 46.3 & 69.9 & 50.4 \\
          NAT-S~\cite{NAT} & Trans & 108 & 809 & 51.9 & 70.4 & 56.2 & 44.9 & 68.2 & 48.6 \\
          GC-ViT-S~\cite{globalvit} & Trans & 108 & 866 & 52.4 & 71.0 & 57.1 & 45.4 & 68.5 & 49.3\\
          DAT-S~\cite{dat} & Trans & 107 & 857 & 52.7 & 71.7 & 57.2 & 45.5 & 69.1 & 49.3 \\
          RMT-B~\cite{fan2023rmt} & Trans & 111 & 852 & 54.5 & 72.8 & 59.0 & 47.2 & 70.5 & 51.4  \\
          \rowcolor{cell}RAVLT-B & Linear & 105 & 851 & \textbf{55.3} & \textbf{73.8} & \textbf{60.1} & \textbf{47.7} & \textbf{71.4} & \textbf{52.1}\\
          \midrule[1pt]
          ConvNeXt-B~\cite{convnext} & CNN & 145 & 964 & 52.7 & 71.3 & 57.2 & 45.6 & 68.9 & 49.5\\
          Swin-B~\cite{SwinTransformer} & Trans & 145 & 982 & 51.9 & 70.9 & 56.5 & 45.0 & 68.4 & 48.7\\
          CSwin-B~\cite{cswin} & Trans & 135 & 1004 & 53.9 & 72.6 & 58.5 & 46.4 & 70.0 & 50.4 \\
          \rowcolor{cell}RAVLT-L & Linear & 152 & 979 & \textbf{55.6} & \textbf{74.1} & \textbf{60.5} & \textbf{48.0} & \textbf{71.8} & \textbf{52.3} \\
          
          \bottomrule[1pt]
    \end{tabular}}
    \vspace{-3mm}
    \caption{Comparison to other backbones using Cascade Mask R-CNN with "$3\times+\mathrm{MS}$" schedule.}
    \vspace{-4mm}
    \label{tab:CASCOCO3x}
\end{table}

\begin{table*}[!ht]
    \setlength{\tabcolsep}{1.25mm}
    \centering
    \scalebox{0.8}{
    \begin{tabular}{c|c|c c|c c c c c c|c c|c c c c c c}
        \toprule[1pt]
        \multirow{2}{*}{Backbone} & Type & \multirow{2}{*}{\makecell{Params\\(M)}} & \multirow{2}{*}{\makecell{FLOPs\\(G)}} & \multicolumn{6}{c|}{Mask R-CNN $1\times$} & \multirow{2}{*}{\makecell{Params\\(M)}} & \multirow{2}{*}{\makecell{FLOPs\\(G)}} & \multicolumn{6}{c}{RetinaNet $1\times$}\\
         & & & & $AP^b$ & $AP^b_{50}$ & $AP^b_{75}$ & $AP^m$ & $AP^m_{50}$ & $AP^m_{75}$ & & & $AP^b$ & $AP^b_{50}$ & $AP^b_{75}$ & $AP^b_S$ & $AP^b_{M}$ & $AP^b_{L}$ \\
         \midrule[1pt]
        PVTv2-B1~\cite{pvtv2} & Trans & 33 & 243 & 41.8 & 54.3 & 45.9 & 38.8 & 61.2 & 41.6 & 23 & 225 & 41.2 & 61.9 & 43.9 & 25.4 & 44.5 & 54.3 \\
        SOFT++-Tiny~\cite{Soft222} & Linear & 32 & 212 & 41.2 & 63.7 & 44.7 & 38.2 & 61.0 & 41.0 & 23 & 200 & 41.9 & 62.7 & 44.7 & 27.8 & 45.4 & 55.6 \\
        FL-PVT-T~\cite{flattentrans} & Linear & 32 & 244 & 38.2 & 61.6 & 41.9 & 37.0 & 57.6 & 39.0 & -- & -- & -- & -- & -- & -- & -- & --\\
        \rowcolor{cell}RAVLT-T & Linear & 33 & 219 & \textbf{47.3} & \textbf{69.1} & \textbf{51.9} & \textbf{42.7} & \textbf{66.2} & \textbf{46.0} & 24 & 201 & \textbf{45.9} & \textbf{67.4} & \textbf{49.4} & \textbf{28.5} & \textbf{50.1} & \textbf{60.8} \\
        \midrule[1pt]
        InternImage-T~\cite{internimage} & CNN & 49 & 270 & 47.2 & 69.0 & 52.1  & 42.5 & 66.1 & 45.8 & -- & -- & -- & -- & -- & -- & -- & -- \\
        CMT-S~\cite{cmt} & Trans & 45 & 249 & 44.6 & 66.8 & 48.9 & 40.7 & 63.9 & 43.4 & 44 & 231 & 44.3 & 65.5 & 47.5 & 27.1 & 48.3 & 59.1 \\
        RMT-S~\cite{fan2023rmt} & Trans & 46 & 262 & 49.0 & 70.8 & 53.9 & 43.9 & 67.8 & 47.4 & 36 & 244 & 47.8 & 69.1 & 51.8 & 32.1 & 51.8 & 63.5 \\
        FL-Swin-T~\cite{flattentrans} & Linear & 49 & 268 & 44.2 & 67.3 & 48.5 & 40.2 & 63.8 & 43.0 & -- & -- & -- & -- & -- & -- & -- & -- \\
        MLLA-T~\cite{MLLA} & Linear & 44 & 255 & 46.8 & 69.5 & 51.5 & 42.1 & 66.4 & 45.0 & -- & -- & -- & -- & -- & -- & -- & -- \\
        \rowcolor{cell}RAVLT-S & Linear & 44 & 262 & \textbf{49.8} & \textbf{71.3} & \textbf{54.5} & \textbf{44.6} & \textbf{68.5} & \textbf{48.2} & 34 & 244 & \textbf{48.3} & \textbf{69.8} & \textbf{52.1} & \textbf{32.7} & \textbf{52.8} & \textbf{63.6} \\
        \midrule[1pt]
        InternImage-S~\cite{internimage} & CNN & 69 & 340 & 47.8 & 69.8 & 52.8 & 43.3 & 67.1 & 46.7 & -- & -- & -- & -- & -- & -- & -- & -- \\
        STViT-B~\cite{stvit} & Trans & 70 & 359 & 49.7 & 71.7 & 54.7 & 44.8 & 68.9 & 48.7 & -- & -- & -- & -- & -- & -- & -- & -- \\
        SOFT++-medium~\cite{Soft222} & Linear & 69 & 342 & 46.6 & 67.8 & 51.2 & 42.0 & 64.8 & 45.2 & 59 & 322 & 44.3 & 64.7 & 47.4 & 29.0 & 48.2 & 59.9 \\
        MLLA-S~\cite{MLLA} & Linear & 63 & 319 & 49.2 & 71.5 & 53.9 & 44.2 & 68.5 & 47.2 & -- & -- & -- & -- & -- & -- & -- & -- \\
        \rowcolor{cell}RAVLT-B & Linear & 67 & 372 & \textbf{51.2} & \textbf{72.7} & \textbf{56.4} & \textbf{45.7} & \textbf{69.9} & \textbf{49.5} & 57 & 353 & \textbf{49.8} & \textbf{71.2} & \textbf{54.0} & \textbf{34.0} & \textbf{54.3} & \textbf{64.9}\\
        \midrule[1pt]
        InternImage-B~\cite{internimage} & CNN & 115 & 501 & 48.8 & 70.9 & 54.0 & 44.0 & 67.8 & 47.4 & -- & -- & -- & -- & -- & -- & -- & -- \\
        Swin-B~\cite{SwinTransformer} & Trans & 107 & 496 & 46.9 & 69.2 & 51.6 & 42.3 & 66.0 & 45.5 & 98 & 477 & 45.0 & 66.4 & 48.3 & 28.4 & 49.1 & 60.6 \\
        RMT-L~\cite{fan2023rmt} & Trans & 114 & 557 & 51.6 & 73.1 & 56.5 & 45.9 & 70.3 & 49.8 & 104 & 537 & 49.4 & 70.6 & 53.1 & 34.2 & 53.9 & 65.2 \\
        MLLA-B~\cite{MLLA} & Linear & 115 & 502 & 50.5 & 72.0 & 55.4 & 45.0 & 69.3 & 48.6 & -- & -- & -- & -- & -- & -- & -- & -- \\
        \rowcolor{cell}RAVLT-L & Linear & 114 & 501 & \textbf{52.3} & \textbf{73.8} & \textbf{57.3} & \textbf{46.4} & \textbf{71.1} & \textbf{50.4} & 104 & 482 & \textbf{50.9} & \textbf{72.2} & \textbf{55.0} & \textbf{34.7} & \textbf{55.7} & \textbf{65.4} \\
        \bottomrule[1pt]
    \end{tabular}}
    \vspace{-3mm}
    \caption{Comparison to other backbones using RetinaNet and Mask R-CNN with ``$1\times$" schedule.}
    \vspace{-3mm}
    \label{tab:COCO1x}
\end{table*}

\paragraph{Results.}We show the comparison results in the Tab.~\ref{tab:COCO3x}, Tab.~\ref{tab:CASCOCO3x}, and Tab.~\ref{tab:COCO1x}. Compared to its competitors, RAVLT achieves excellent results across every detection framework. Specifically, with the framework of Cascade Mask R-CNN and ``$3\times + MS$" schedule, RAVLT-B achieves \textbf{55.3$AP^b$} and \textbf{47.7$AP^m$}. The result even surpasses CSwin-B, which is much larger than RAVLT-B. 

\begin{table}[t]
    \centering
    \setlength{\tabcolsep}{0.45mm}
    \scalebox{0.8}{
    \begin{tabular}{c|c|c c c|c c c}
    \toprule[1pt]
    \multirow{3}{*}{Model} & \multirow{3}{*}{Type} & \multicolumn{3}{c|}{Semantic FPN 80K} & \multicolumn{3}{c}{Upernet 160K} \\
    & & \makecell{Params\\(M)} & \makecell{FLOPs\\(G)} & \makecell{mIoU\\(\%)} & \makecell{Params\\(M)} & \makecell{FLOPs\\(G)} & \makecell{mIoU$_{ss}$\\(\%)} \\
    \midrule[0.5pt]
    VAN-B1~\cite{VAN} & CNN & 18 & 140 & 42.9 & -- & -- & -- \\
    PVTv2-B1~\cite{pvtv2} & Trans & 18 & 136 & 42.5 & -- & -- & -- \\
    RMT-T~\cite{fan2023rmt} & Trans & 17 & 136 & 46.4 & -- & -- & -- \\
    FL-PVT-T~\cite{flattentrans} & Linear & 16 & 169 & 37.2 & -- & -- & -- \\
    \rowcolor{cell}RAVLT-T & Linear & 18 & 136 & \textbf{47.9} & 44 & 893 & \textbf{49.3} \\
    \midrule[0.5pt]
    MogaNet-S~\cite{iclr2024MogaNet} & CNN & 29 & 189 & 47.7 & 55 & 946 & 49.2 \\
    StructViT-S~\cite{structvit} & Trans & 26 & 271 & 46.9 & -- & -- & -- \\
    SMT-S~\cite{SMT} & Trans & -- & -- & -- & 50 & 935 & 49.2 \\
    SGFormer-S~\cite{sgformer} & Trans & 25 & 205 & 49.0 & 53 & 989 & 49.9 \\
    RMT-S~\cite{fan2023rmt} & Trans & 30 & 180 & 49.4 & 56 & 937 & 49.8 \\
    Fl-Swin-T~\cite{flattentrans} & Linear & -- & -- & -- & 60 & 946 & 44.8 \\
    \rowcolor{cell}RAVLT-S & Linear & 28 & 180 & \textbf{49.5} & 55 & 937 & \textbf{50.7} \\
    \midrule[0.5pt]
    MogaNet-B~\cite{iclr2024MogaNet} & CNN & -- & -- & -- & 74 & 1050 & 50.1 \\
    InterImage-S~\cite{internimage} & CNN & -- & -- & -- & 80 & 1017 & 50.2 \\
    DAT-S~\cite{dat} & Trans & 53 & 320 & 46.1 & 81 & 1079 & 48.3 \\
    StructViT-B~\cite{structvit} & Trans & 54 & 529 & 48.5 & -- & -- & -- \\
    RMT-B~\cite{fan2023rmt} & Trans & 57 & 294 & 50.4 & 83 & 1051 & 52.0 \\
    FL-Swin-S~\cite{flattentrans} & Linear & -- & -- & -- & 82 & 1038 & 48.1 \\
    \rowcolor{cell}RAVLT-B & Linear & 51 & 292 & \textbf{51.9} & 77 & 1050 & \textbf{52.5} \\
    \midrule[0.5pt]
    MogaNet-L~\cite{iclr2024MogaNet} & CNN & -- & -- & -- & 113 &1176 & 50.9 \\
    CSWin-B~\cite{cswin} & Trans & 81 & 464 & 49.9 & 109 & 1222 & 51.1 \\
    SGFormer-B~\cite{sgformer} & Trans & 81 & 475 & 50.6 & 109 & 1304 & 52.0 \\
    RMT-L~\cite{fan2023rmt} & Trans & 98 & 482 & 51.4 & 125 & 1241 & 52.8 \\
    MLLA-B~\cite{MLLA} & Linear & -- & -- & -- & 128 & 1183 & 51.9 \\
    \rowcolor{cell}RAVLT-L & Linear & 98 & 424 & \textbf{52.6} & 125 & 1182 & \textbf{53.2} \\
    \bottomrule[1pt]

    \end{tabular}}
    \vspace{-3mm}
    \caption{ Comparison with the state-of-the-art on ADE20K.}
    \vspace{-7mm}
    \label{tab:seg}
\end{table}

\subsection{Semantic Segmentation}
\paragraph{Settings.}We adopt the Semantic FPN~\cite{semanticfpn} and UperNet~\cite{upernet} based on MMSegmentation~\cite{mmsegmentation}, apply RAVLTs which are pretrained on ImageNet-1K as backbone. We use the same setting of PVT~\cite{pvt} to train the Semantic FPN, and follow the default settings in Swin~\cite{SwinTransformer} to train the UperNet.

\paragraph{Results.}We show the segmentation results in the Tab.~\ref{tab:seg}. RAVLT surpasses its competitors across various frameworks. For example, with the framework of Semantic FPN, RAVLT-B achieves \textbf{51.9}mIoU. The result even surpasses larger RMT-L. With the framework of UperNet, RAVLT can achieve the mIoU of \textbf{53.2}. 

\subsection{Comparison with Other Linear Attention}
For a fair comparison with other linear attention mechanisms, similar to FLatten-Transformer~\cite{flattentrans}, we use the DeiT-T~\cite{deit}, Swin-T, and Swin-S~\cite{SwinTransformer} network architectures, replacing their attention mechanisms with linear attention mechanisms. As shown in Tab.~\ref{tab:strcomp}, our RALA significantly outperforms other linear attention mechanisms and also exceeds the performance of Softmax attention.

\begin{table}[ht]
    \centering
    \setlength{\tabcolsep}{1.1mm}
    \scalebox{0.85}{
    \begin{tabular}{c|c c|c}
    \toprule[1pt]         
         Linear Attention & Params(M) & FLOPs(G) & Top1-acc(\%)\\
         \midrule[0.5pt]
         \multicolumn{4}{c}{Comparison on DeiT-T Setting} \\
         \midrule[0.125pt]
         DeiT-T~\cite{deit} & 6 & 1.1 & 72.2 \\
         Hydra Attn~\cite{hydraattn} & 6 & 1.1 & 68.3 \\
         Efficient Attn~\cite{efficientattn} & 6 & 1.1 & 70.2 \\
         Linear Angular Attn~\cite{you2023castling} & 6 & 1.1 & 70.8 \\
         Enhanced Linear Attn~\cite{efficientvit} & 6 & 1.1 & 72.9 \\
         Focused Linear Attn~\cite{flattentrans} & 6 & 1.1 & 74.1 \\
         \rowcolor{cell}Ours & 6 & 1.1 & \textbf{75.1} \\
         \midrule[0.5pt]
         \multicolumn{4}{c}{Comparison on Swin-T Setting} \\
         \midrule[0.125pt]
         Swin-T~\cite{SwinTransformer} & 29 & 4.5 & 81.3 \\
         Hydra Attn~\cite{hydraattn} & 29 & 4.5 & 80.7 \\
         Efficient Attn~\cite{efficientattn} & 29 & 4.5 & 81.0 \\
         Linear Angular Attn~\cite{you2023castling} & 29 & 4.5 & 79.4 \\
         Enhanced Linear Attn~\cite{efficientvit} & 29 & 4.5 & 81.8 \\
         Focused Linear Attn~\cite{flattentrans} & 29 & 4.5 & 82.1 \\
         \rowcolor{cell}Ours & 29 & 4.5 & \textbf{83.4} \\
         \midrule[0.5pt]
         \multicolumn{4}{c}{Comparison on Swin-S Setting} \\
         \midrule[0.125pt]
         Swin-S~\cite{SwinTransformer} & 50 & 8.7 & 83.0 \\
         Focused Linear Attn~\cite{flattentrans} & 51 & 8.7 & 83.5 \\
         \rowcolor{cell}Ours & 50 & 8.7 & \textbf{85.0} \\
         \bottomrule[1pt]
    \end{tabular}}
    \vspace{-3mm}
    \caption{Comparison of different linear attention based on DeiT-T, Swin-T, and Swin-S. RALA surpasses others by a large margin.}
    \vspace{-5mm}
    \label{tab:strcomp}
\end{table}

\subsection{Ablation Study}

\paragraph{Augmentation of the KV buffer.}We first ablate the effect of augmenting the KV buffer on the model. As shown in Tab.~\ref{tab:ab}, augmenting the KV buffer significantly increases its rank and enhances feature diversity, thereby leading to substantial improvements in model performance. This enhancement is particularly evident in downstream tasks such as object detection.

\vspace{-4mm}

\paragraph{Augmentation of the output features.}Since we use a linear projection to serve as $\phi(.)$, ablating the augmentation of the output feature reduces the model's parameter count. To ensure a fair comparison, we increase the expansion ratio of the FFN to maintain the model's parameter count and FLOPs at the same level. As shown in Tab.~\ref{tab:ab}, the introduction of augmentation effectively increases the rank of the output features, thereby enhancing model's performance.

\vspace{-4mm}

\paragraph{Conditional Position Encoding.}We use CPE~\cite{CPVT} to provide the positional information to our model. As shown in Tab.~\ref{tab:ab}, CPE also improves the model's performance to some extent across various tasks. This indicates that CPE provides effective positional information for RAVLT.

\vspace{-4mm}

\paragraph{Selection of $\phi(.)$ and $\kappa(.)$.}We compare the choices of functions $\phi(.)$ (linear projection, identity map, and tanh) and $\kappa(.)$ (${\rm Elu}(.)+1$ and ${\rm ReLU}(.)$), and experiments show that the model's performance improves with the inclusion of $\phi(.)$, regardless of the specific choice of it. This conclusion also holds for $\kappa(.)$, as shown in Tab.~\ref{tab:ab}. \textbf{This indicates that the performance improvements stem from our RALA's augmentation of matrix's rank, rather than the selection of specific functions}.

\begin{table}[ht]
    \centering
    \setlength{\tabcolsep}{0.25mm}
    \scalebox{0.82}{
    \begin{tabular}{c|c c|c c c c}
    \toprule[1pt]
         & \makecell{Params} & \makecell{FLOPs} & \makecell{Top1} & $AP^b$ & $AP^m$ & mIoU\\
         \midrule[0.5pt]
         & 15 & 2.4 & 82.8 & 47.3 & 42.7 & 47.9 \\
         \midrule[0.5pt]
         w/o KV aug & 15 & 2.4 & 82.1(\textcolor{red}{-0.7}) & 43.7(\textcolor{red}{-3.6}) & 39.0(\textcolor{red}{-3.7}) & 43.6(\textcolor{red}{-4.3})\\
         w/o out aug & 15 & 2.4 & 82.5(\textcolor{red}{-0.3}) & 46.3(\textcolor{red}{-1.0}) & 41.6(\textcolor{red}{-1.1}) & 47.0(\textcolor{red}{-0.9})\\
         w/o CPE & 15 & 2.4 & 82.7(\textcolor{red}{-0.1}) & 47.0(\textcolor{red}{-0.3}) & 42.5(\textcolor{red}{-0.2}) & 47.6(\textcolor{red}{-0.3})\\
         \midrule[0.5pt]
         \makecell{$\phi$: proj$\xrightarrow{}$\\identity} & 15 & 2.4 & 82.8(\textcolor{blue}{-0.0}) & 47.2(\textcolor{red}{-0.1}) & 42.5(\textcolor{red}{-0.2}) & 48.1(\textcolor{blue}{+0.2}) \\
         \makecell{$\phi$: proj$\xrightarrow{}$\\tanh} & 15 & 2.4 & 82.7(\textcolor{red}{-0.1}) & 47.2(\textcolor{red}{-0.1}) & 42.7(\textcolor{blue}{-0.0}) & 47.7(\textcolor{red}{-0.2}) \\
         \makecell{$\kappa$: ${\rm ELU}+1$\\$\xrightarrow{}{\rm ReLU}$} & 15 & 2.4 & 82.8(\textcolor{blue}{-0.0}) & 47.3(\textcolor{red}{-0.0}) & 42.8(\textcolor{blue}{+0.1}) & 47.7(\textcolor{red}{-0.2}) \\
         \makecell{$\kappa$: ${\rm ELU}+1$\\$\xrightarrow{}{\rm Softmax}$} & 15 & 2.4 & 82.7(\textcolor{red}{-0.1}) & 47.1(\textcolor{red}{-0.2}) & 42.7(\textcolor{blue}{+0.0}) & 47.8(\textcolor{red}{-0.1}) \\
         \bottomrule[1pt]

    \end{tabular}}
    \vspace{-3mm}
    \caption{Ablation on RAVLT-T. The results show that performance improvements stem from our RALA's augmentation of matrix's rank, rather than the selection of specific functions.}
    \vspace{-3mm}
    \label{tab:ab}
\end{table}

\vspace{-4mm}

\begin{figure}
    \centering
    \includegraphics[width=0.98\linewidth]{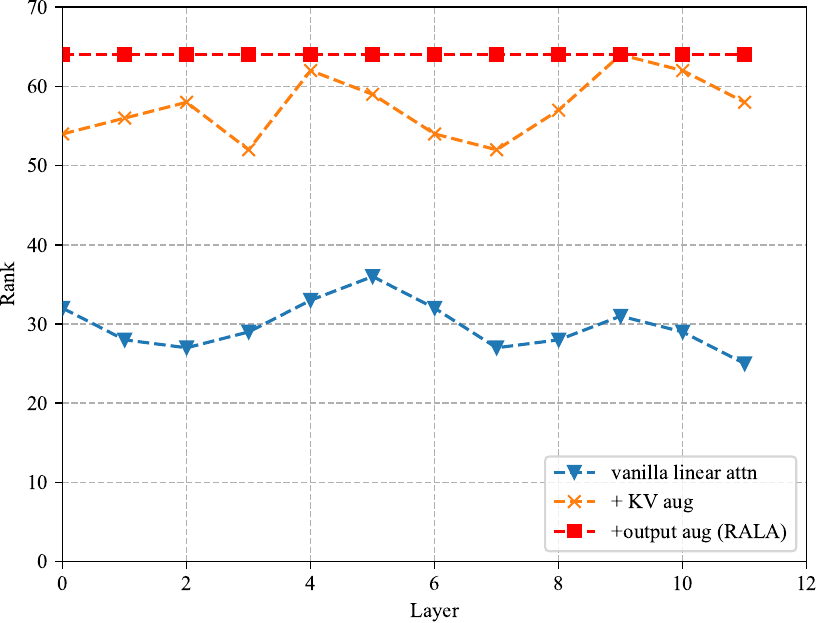}
    \vspace{-3mm}
    \caption{Analysis of the rank across different model layers. RALA's augmentation of the KV buffer and output feature effectively increases the rank of the output features at each layer.}
    \vspace{-5mm}
    \label{fig:nanrank}
\end{figure}

\paragraph{Rank analysis of different layers.}Under the DeiT-T setting, we analyze the rank of different layers of our model (12 layers in total, with each layer's feature matrix having a shape of $N=196$, $d=64$, and a full rank of 64). As shown in Fig.~\ref{fig:nanrank}, RALA effectively increases the rank of the model's output features, thereby enhancing its performance.

\section{Conclusion}

In this paper, we investigate the causes of performance degradation in linear attention and conclude that the low rank of its output features is the primary factor. To address this issue, we propose Rank-Augmented Linear Attention (RALA). This method enhances two critical computation steps in linear attention—the calculation of the KV buffer and the computation of the output features—by employing two transformations. These augmentations effectively increase the rank of the matrices, thereby mitigating the low-rank problem in linear attention. Building upon RALA, we develop the Rank-Augmented Vision Linear Transformer (RAVLT), a novel architecture based on linear attention. RAVLT outperforms many existing Transformers, demonstrating significant potential for practical applications.
\section{Acknowledgements}
This work is partially funded by Beijing Natural Science Foundation (4252054), Youth Innovation Promotion Association CAS(Grant No.2022132), Beijing Nova Program(20230484276), and CCF-Kuaishou Large Model Explorer Fund (NO. CCF-KuaiShou 2024005).
\clearpage
\setcounter{page}{1}
\maketitlesupplementary
\begin{appendices}

\section{More Analysis About $\alpha$}
In this section, we further analyze the rank-boosting effect of $\alpha$ on the KV buffer and conduct additional ablation experiments on $\alpha$ to validate the rationale behind our design.
\paragraph{An analysis of the rank-boosting effect.}Assume in $\kappa(K_j^T)V_j, j\in[0, N]$ that the two components $\kappa(K_1^T)V_1$ and $\kappa(K_2^T)V_2$ are linearly correlated. That means:
\begin{equation}
    \kappa(K_1^T)V_1 = \lambda\kappa(K_2^T)V_2.
\end{equation}
In general linear attention mechanisms, a portion of the KV buffer formed by the sum of the two can be expressed as:
\begin{equation}
\label{eq:s1}
    \kappa(K_1^T)V_1+\kappa(K_2^T)V_2 = (1+\lambda)\kappa(K_2^T)V_2
\end{equation}
After introducing $\alpha$ as a modulation coefficient for each term, their sum becomes:
\begin{equation}
\label{eq:s2}
    \begin{aligned}
        \alpha_1\kappa(K_1^T)V_1+\alpha_2\kappa(K_2^T)V_2&=\lambda\alpha_1\kappa(K_2^T)V_2+\alpha_2\kappa(K_2^T)V_2\\
        &=\alpha_2(1+\frac{\lambda\alpha_1}{\alpha_2})\kappa(K_2^T)V_2\\
    \end{aligned}
\end{equation}
Since in our setup the value of $\alpha_1$ and $\alpha_2$ are calculated based on the attention scores between global query and keys, it changes with the input samples and the training process. This results in more varied coefficients, meaning that the new matrix composed of these two linearly correlated matrices has a broader range of possible values and greater flexibility. This makes the KV buffer $\sum_{j=1}^N\alpha_j\kappa(K_j^T)V_j$ more likely to achieve a full-rank state. 

As previously mentioned (Eq.~\ref{eq:s1} and Eq.~\ref{eq:s2}), two linearly correlated matrices can be summed to form a single matrix with coefficients. Therefore, we consider here only the case where all matrices are linearly independent. Consider the equation:
\begin{equation}
    \sum_{j=1}^Nc_j\kappa(K_j^T)V_j=C
\end{equation}
Here, $C\in \mathbb{R}^{d\times d}$ is a full-rank matrix. Since all $\kappa(K_j^T)V_j$ are linearly independent, this equation has a \textbf{unique solution or no solution}. After decomposing $c_j$ into $d_j\alpha_j$, the original equation becomes:
\begin{equation}
    \sum_{j=1}^Nd_j\alpha_j\kappa(K_j^T)V_j=C
\end{equation}
Due to the presence of different scalars, the solutions can be multiple, giving the matrix representation a broader scope. This implies that there are more solutions that enable the KV buffer to achieve a full-rank state.

\paragraph{More choices about $\alpha$.}We choose the attention scores between the keys and the global query as the value for $\alpha$, and here we experiment with different selections. When $\alpha$ is set as a learnable vector, its fixed shape of $1\times N$ makes it challenging to apply to higher resolution tasks such as object detection. Therefore, we only evaluate its performance on image classification tasks. 
\begin{table}[h]
    \centering
    \begin{tabular}{c|c c|c}
         \toprule[1pt]
         Model & Params(M) & FLOPs(G) & Acc(\%)\\
         \midrule[0.5pt]
         DeiT-T & 6 & 1.1 & 72.2 \\
         \midrule[0.5pt]
         attn score & 6 & 1.1 & 75.1 \\
         learnable & 6 & 1.1 & 73.8(\textcolor{red}{-1.3}) \\ 
         \bottomrule[1pt]
    \end{tabular}
    \vspace{-3mm}
    \caption{More choice about $\alpha$.}
    \vspace{-3mm}
    \label{tab:abalpha}
\end{table}
We conduct experiments using the DeiT-T~\cite{deit} configuration, and the results are shown in Tab.~\ref{tab:abalpha}. While the learnable $\alpha$ still significantly enhances the model's performance, its effectiveness is not as impressive as the $\alpha$ based on the attention score.

\section{More Analysis About $\phi(.)$}

\paragraph{An analysis of the rank-boosting effect.}Consider two matrices $A$ and $B$, $A,B\in \mathbb{R}^{m\times n}$, and represent these matrices as a linear combination of rank-one matrices. That is:
\begin{equation}
    \begin{aligned}
    A=\sum_{i=1}^ru_iv_i^T, \quad B=\sum_{j=1}^sx_jy_j^T\\
    \end{aligned}
\end{equation}
where $r={\rm Rank}(A)$, $s={\rm Rank}(B)$, $u_i, x_j \in \mathbb{R}^{m\times 1}$, $v_i, y_j \in \mathbb{R}^{n\times 1}$. Consider the Hardmard product:
\begin{equation}
    (A\odot B)_{ij}=A_{ij}\odot B_{ij}
\end{equation}
we can rewrite it as:
\begin{equation}
\begin{aligned}
    A\odot B &= (\sum_{i=1}^ru_iv_i^T)\odot (\sum_{j=1}^sx_jy_j^T) \\
    &=\sum_{i=1}^r \sum_{j=1}^s (u_i\odot x_j)(v_i\odot y_j)^T
\end{aligned}
\end{equation}
This expression shows that $A\odot B$ can be viewed as a linear combination of $rs$ rank-one matrices. Therefore, the total rank of $A\odot B$ does not exceed the number of matrices, which is:
\begin{equation}
    {\rm Rank}(A\odot B) \leq rs={\rm Rank}(A)\times {\rm Rank}(B)
\end{equation}
The above expression effectively demonstrates that when the ranks of the two matrices, $r$ and $s$, are both small, the Hadamard product can effectively raise the upper bound of the matrix rank. Therefore, when augmenting the rank of the output features matrix, the choice of $\phi(.)$ becomes less important; what matters is the use of the Hadamard product. \textcolor{red}{This is consistent with the conclusions we obtained from the ablation experiments in the main text.} Regardless of what $\phi(.)$ is, the presence of $\phi(.)$ will always enhance the rank of the matrix.

\section{Experimental Details}

\paragraph{Image Classification.}
We adopt the training strategy proposed in DeiT~\cite{deit} with the only supervision is classification loss. Specifically, all models are trained from scratch for 300 epochs with the input resolution of $224\times 224$. Adam is used with a cosine decay learning rate scheduler and 5 epochs of linear warm-up. The initial learning rate,  weight decay, and  batch-size are set to  0.001, 0.05, and 1024, respectively. We apply the same data augmentation and regularization as DeiT~\cite{deit} (RandAugment \cite{randomaugment} (randm9-mstd0.5-inc1) , Mixup \cite{mixup} (prob = 0.8), CutMix \cite{cutmix} (prob = 1.0), Random Erasing (prob = 0.25)).

\paragraph{Object Detection and Instance Segmentation.}
We apply RetinaNet~\cite{retinanet}, Mask-RCNN~\cite{maskrcnn}, and Cascaded Mask R-CNN~\cite{cai18cascadercnn} as the frameworks based on the MMDetection \cite{mmdetection} to evaluate our models. The models are trained under ``1 $\times$" (12 training epochs) and ``3 $\times$ +MS" (36 training epochs with multi-scale training) settings. For the ``1 $\times$" setting, images are resized to the shorter side of 800 pixels while the longer side is within 1333 pixels. For the ``3 $\times$ +MS", multi-scale training strategy is applied to randomly resize the shorter side between 480 to 800 pixels. We use the initial learning rate of 1e-4. For RetinaNet, we set the weight decay to 1e-4. For Mask-RCNN and Cascaded Mask R-CNN, we set it to 5e-2.

\paragraph{Semantic Segmentation.}we implement UperNet~\cite{upernet} and SemanticFPN~\cite{semanticfpn} based on MMSegmentation~\cite{mmsegmentation} to validate the models. For UperNet, we follow the previous setting~\cite{SwinTransformer} and train the model for 160k iterations with the input size of $512\times 512$. For SemanticFPN, we also use the input resolution of $512\times512$ but train the models for 80k iterations.

\end{appendices}
{
    \small
    \bibliographystyle{ieeenat_fullname}
    \bibliography{main}
}


\end{document}